\documentclass[letterpaper, 10 pt, conference]{ieeeconf}  % Comment this line out if you need a4paper

\IEEEoverridecommandlockouts                              % This command is only needed if 
\usepackage{graphicx} % for pdf, bitmapped graphics files
\graphicspath{{images/}}
\DeclareGraphicsExtensions{.pdf,.jpg,.png}
\usepackage[intlimits]{amsmath} % assumes amsmath package installed
\usepackage{amssymb}  % assumes amsmath package installed
\usepackage{amsmath}  % assumes amsmath package installed
\usepackage{url}
\usepackage{subcaption}
\usepackage{wrapfig}
\usepackage[autostyle]{csquotes}
\usepackage[font={small}]{caption}
\usepackage[table,xcdraw]{xcolor}
\usepackage{algorithm}
\usepackage[noend]{algpseudocode}
\usepackage{booktabs}

\algrenewcommand\algorithmicindent{1.0em}%

\makeatletter
\renewcommand{\ALG@beginalgorithmic}{\small}
\makeatother

\def\secref#1{Sec.~\ref{#1}}
\def\figref#1{Fig.~\ref{#1}}

\def\eqref#1{Eq.~(\ref{#1})}

\newcommand{\etal}{~\textit{et~al.\;}}

\newcommand{\ie}{i.\,e., }

\newcommand{\eg}{e.\,g., }

\title{\LARGE \bf Fast-Replanning Motion Control for Non-Holonomic Vehicles\\ with Aborting A* }

\author{Marcell Missura \and Arindam Roychoudhury \and Maren Bennewitz% <-this % stops a space
\thanks{All authors are with the Humanoid Robots Lab, University of Bonn, Germany. Contact:
        {\tt\small missura@cs.uni-bonn.de}}%
}

\begin{document}

\maketitle
\thispagestyle{empty}
\pagestyle{empty}

%%%%%%%%%%%%%%%%%%%%%%%%%%%%%%%%%%%%%%%%%%%%%%%%%%%%%%%%%%%%%%%%%%%%%%%%%%%%%%%%
\begin{abstract}

Autonomously driving vehicles must be able to navigate in dynamic and unpredictable environments in a collision-free manner. So far, this has only been partially achieved in driverless cars and warehouse installations where marked structures such as roads, lanes, and traffic signs simplify the motion planning and collision avoidance problem. We are presenting a new control approach for car-like vehicles that is based on an unprecedentedly fast-paced A* implementation that allows the control cycle to run at a frequency of 30~Hz. This frequency enables us to place our A* algorithm as a low-level replanning controller that is well suited for navigation and collision avoidance in virtually any dynamic environment. Due to an efficient heuristic consisting of rotate-translate-rotate motions laid out along the shortest path to the target, our Short-Term Aborting A* (STAA*) converges fast and can be aborted early in order to guarantee a high and steady control rate. While our STAA* expands states along the shortest path, it takes care of collision checking with the environment including predicted states of moving obstacles, and returns the best solution found when the computation time runs out. Despite the bounded computation time, our STAA* does not get trapped in corners due to the following of the shortest path. In simulated and real-robot experiments, we demonstrate that our control approach eliminates collisions almost entirely and is superior to an improved version of the Dynamic Window Approach with predictive collision avoidance capabilities~\cite{Missura:DynamicDWA}.

\end{abstract}

%%%%%%%%%%%%%%%%%%%%%%%%%%%%%%%%%%%%%%%%%%%%%%%%%%%%%%%%%%%%%%%%%%%%%%%%%%%%%%%%
\section{INTRODUCTION}

Dynamic environments as for example pedestrian zones or the corridors of an office building, where humans and other robots move freely without following traffic rules, pose a particulalry hard challenge for the design of reliable motion controllers that are effective at avoiding collisions and reaching goal locations. Such environments are unpredictable to a large extent and reaction by fast replanning is the only way of coping with unforeseen events such as the change of direction of a nearby agent, or the sudden appearance of an obstacle in the sensor range that is already on collision course. Fast replanning, however, also implies short computation times that do not suffice for the computation of a detailed plan all the way to the target. The repeated execution of an incomplete plan can, however, result in eternal osscillations or imprisonment in local minima as it is the case with the original Dynamic Window Approach (DWA)~\cite{DWA}. Due to this reason, two-tiered navigation controllers emerged that compute a high-level plan at a slow rate all the way to the target (\eg the shortest path), and follow the plan with a short-sighted, low-level controller that cares for the path following and collision avoidance at a high frequency. The work of Willow Garage~\cite{OfficeMarathon} constitutes an extensive test of this control paradigm where a vanilla A*~grid search was used to find the shortest path in an occupancy map and an intermediate target along this path serves as a local goal for the DWA that follows the path and avoids collisions.

\begin{figure}[t]
\vspace{2.5mm}
      \centering
      \frame{\includegraphics[width=0.98\columnwidth]{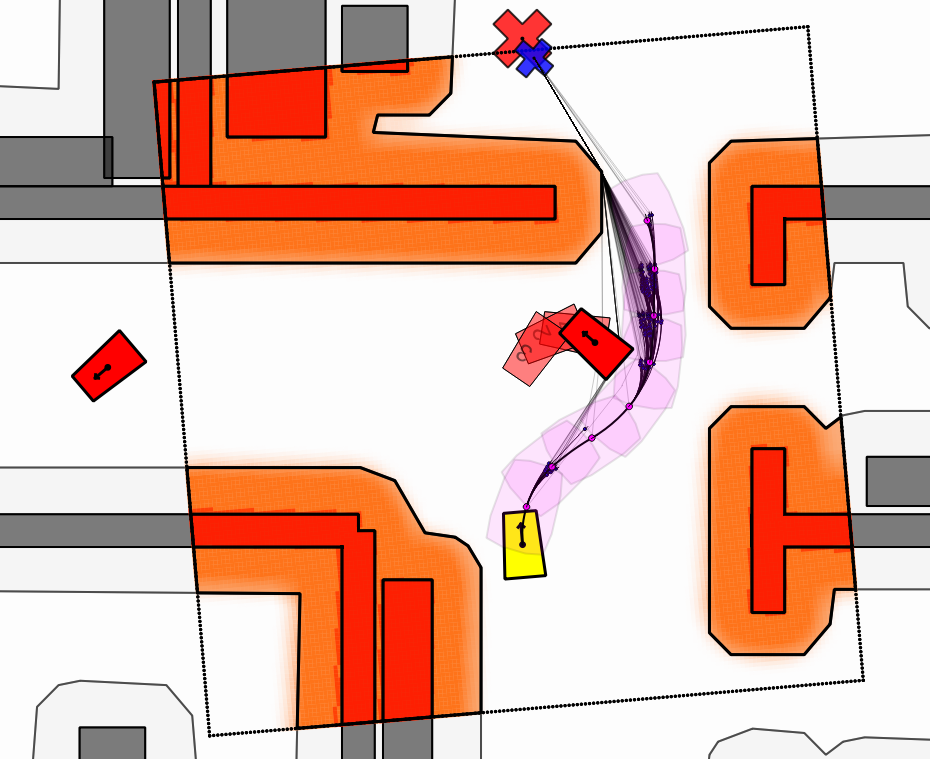}}
      \caption{Our Short-Term Aborting A* controller computing a trajectory for the yellow agent while taking into account predicted states of another moving agent~(fading red). Our controller replans at a rate of 30\,Hz and aborts the search early after a bounded time. The planning is performed in a local map around the yellow agent that contains expanded polygons~(black outline) and a grid representation~(orange and red areas) for a costmap and collision checking. The blue cross at the upper boundary of the local map is an intermediate target extracted from the global path, which is planned in the underlying polygonal global map shown in grey. The shortest paths indicated by thin lines between the search graph and the target are used for the heuristic of the A*.}
      \label{fig:everything}
\vspace{-6mm}
\end{figure}

Our control framework is two-tiered in the same fashion. We use the shortest path as a navigation guide and extract from it a medium-term target for the lower layer. On the lower layer, we use a bounded-time A* implementation that enables us to recompute the control cycle at a frequency of 30\,Hz. 
The A* algorithm applies a discrete set of acceleration commands to the current state of the non-holonomic agent for a short time and predicts their outcome including the expected motion of other moving agents. From those predicted states, it chooses the most promising one according to collision considerations and a heuristic cost function. It repeatedly expands the most promising state until either the target state has been reached, or the allowed computation time has run out. The strength of our A* implementation is an obstacle-aware heuristic function that lays out rotate-translate-rotate motions along the shortest path. Expanding along the shortest path makes sure that aborting the search early does not trap the agent in a local minimum. We have already applied this heuristic to footstep planning \cite{Missura:FastFootstepPlanning} and achieved a high replanning rate with an A* search. In this contribution, we transfer the same principle to non-holonomic vehicles and obtain a similar performance.

Furthermore, our system no longer relies on memory-consuming grid maps that can become prohibitively large. Instead, we assume a \emph{geometric map} of the environment where the obstacles are represented as a set of polygons. We use only a small local grid around the agent for processing sensory input and speeding up collision checking. An overview of our system is illustrated in Figure~\ref{fig:everything}.

\section{RELATED WORK}

%Early versions of navigational software relied on the force field method \cite{PotentialField} \cite{PotentialField2}. Obstacles in the environment exert a repulsive force on the agent while the target is an attractor. However, the purely reactive nature of this approach renders foresighted motion out of reach. Without the guidance of a path, it is also bound to get trapped in a local minimum.

Despite being an aging concept, the Dynamic Window Approach~\cite{DWA} (DWA) is still the most dominant approach to collision-avoiding motion control to date. The DWA uses circular arcs to approximate and predict the otherwise complex trajectories of non-holonomic vehicles \cite{Missura:wheeled}. After systematic sampling of such arcs, the best of them is elected according to an objective function that evaluates obstacle clearance and progress towards the target. The elected arc then yields the next velocity command that is sent to the robot. Due to its limited planning horizon, the DWA is susceptible to getting stuck in local minima. To mitigate this issue, the Global Dynamic Window Approach \cite{GlobalDWA} uses a precomputed Dijkstra heuristic to guide the DWA to the goal. In our previous work, we extended the DWA with predictive collision-avoidance capabilities with regards to moving obstacles~\cite{Missura:DynamicDWA} and guide the DWA by following a short-term goal extracted from the shortest path to the target. %Our new approach borrows the concept of approximating non-holonomic trajectories with arcs, but uses an informed heuristic search of arbitrary depth to plan ahead instead of the limited horizon of the DWA. 

Safe Interval Path Planning (SIPP) \cite{SIPP} is similar to our work in the sense that it plans motion considering moving obstacles. It is a clever A* planning algorithm in a grid world where the time dimension is replaced by a much more compact number of safe intervals during which a certain configuration is known to remain collision-free. This reduction and a precomputed map of shortest paths allow the system to compute a coarse motion plan in the matter of a few hundred milliseconds. Our system is an order of magnitude faster and contrary to SIPP, guarantees the completion of a suboptimal plan within bounded time. There are time-bounded versions of A* that interleave planning and execution such as TBA* \cite{TBAS}, where the number of A* expansions is limited between actions taken by the robot and the same search is continued over the course of multiple actions until the optimal solution is found. However, discovering a shorter path enroute can result in the agent having to backtrack. RTAA* \cite{RTAA} and LRTA* \cite{LRTAS} limit the depth of the search and the agent updates the heuristic of visited cells as it moves. Apart from the drawbacks of discretization, all grid-based solutions suffer from the fact that especially memory requirements, but also runtime, increase with the size of the map. Our system is embedded into continuous, geometric space where not the size, but only the local complexity of the environment influences the runtime of a search. %Expecting rapid changes in the environment, we perform pure replanning from scratch and rebuild a local model in every frame.%
In continuous space, we make no sacrifice in smoothness and can model the acceleration-controlled dynamics of the vehicle more accurately using the unicycle model as opposed to the coarsely discretized chess figure models that emerge from grid-based planning.

Other non time-bounded, but dynamic flavours of the A* algorithm include for example D*-Lite \cite{DStarLite}, where a Dijkstra table is maintained by updating only the necessary portions of it while the robot is driving and discovering changes in its environment. Likachev\etal\cite{Likachev} presented a highly evolved controller for cars based on a combination of D*-Lite and ARA* \cite{ARAStar} where an inflated heuristic is used to compute a suboptimal plan in shorter time that can be refined with a less inflated heuristic as long as deliberation time allows. A multiresolutionally discretized lattice of the state space and an obstacle-aware heuristic were used to speed up the search along with the reduction of the state space by ignoring curvature. The planner achieved a replanning frequency around 1\,Hz with highly varying runtimes and a subcontroller running at 10\,Hz had to be used to follow the computed plan. Since both D*-Lite and the non-holonomic A* in \cite{Likachev} plan from the target to the start, premature abortion is not possible.

A different class of motion planners attempts to refine an A* path to a dynamically feasible motion trajectory on the fly. Quintic splines \cite{Lau}, B-splines \cite{Usenko}, Bezier curves~\cite{Sterling}, or elastic bands \cite{Kinodynamic} are used in a manner of online optimization. While the path to begin with is collision-free, the refined trajectory is not necessary so, especially not if the obstacles move between updates.

Machine learning has been extensively experimented with in order to learn a collision-avoiding navigation controller. For example, Pfeiffer\etal~\cite{Siegwart} propposed to train an end-to-end deep network motion planner for small, static maps using the raw sensor data as input, the motor controls as output, and a ROS path planner as the expert teacher. Chen, Everett\etal\cite{Everett1,Everett2} trained a deep network to navigate among moving obstacles. In order to process an undetermined number of obstacles, an LSTM layer was used to convert an arbitrary number of observed agents to a fixed size input for the network. The obstacles were represented as circular objects encoded by their position, velocity, and radius. This setup is not well suited to encode the large, non-convex structures of indoor environments. Kretzschmar\etal\cite{Kretzschmar} used inverse reinforcement learning to identify the parameters of human trajectories that were modeled by cubic splines to achieve a predictive system with human-like behavior. % In their target environments, these systems achieved results comparable to or superior to engineered approaches. 

A related geometric approach to obstacle avoidance is based on the concept of Velocity Obstacles~\cite{VO}. Velocity Obstacles are polygonal regions in velocity space that describe the set of velocities an agent A is not permitted to use, otherwise it will eventually collide with agent B, if both agents continue to travel with the same velocity. Assuming two or more agents are using the same control principle and choose a velocity for the next time step outside of their respective velocity obstacles in a reciprocal fashion \cite{RVO}, a simple linear program can be used to guarantee collision-free motion for n-bodies~\cite{NRVO} as long as the modeling assumptions hold. Acceleration-velocity obstacles \cite{AVO} and non-holonomic control \cite{VO3} have also been investigated. Velocity Obstacles assume convex polygonal obstacles that use the same controller and travel at constant velocity. Our concept can also deal with the non-convex polygons of arbitrary size that make up most of the environment and does not make the assumption of a certain type of controller being used by the other agents. Furthermore, as it is possible to enter and leave a velocity obstacle before the collision actually occurs, our A* approach provides far more precise collision checking than the Velocity Obstacle concept.

\section{MOTION CONTROL FRAMEWORK}

%\begin{figure*}[t]
%      \centering
%      \begin{subfigure}[b]{.27\linewidth}
%      \frame{\includegraphics[height=5cm]{images/pipeline5.png}}
%      \caption{Static grid and polygon maps extracted from the map.}
%      \end{subfigure}
%      \hspace{1cm}
%      \begin{subfigure}[b]{.27\linewidth}
%      \frame{\includegraphics[height=5cm]{images/dynamicpolygons.png}}
%      \caption{Dynamic polygons with estimated velocity vectors (length of arrows).}
%      \end{subfigure}
%      \hspace{1cm}
%      \begin{subfigure}[b]{.27\linewidth}
%      \frame{\includegraphics[height=5cm]{images/predictedpathplanning.png}}
%      \caption{Static and dynamic polygons being used for path planning.}
%      \end{subfigure}
%      \caption{Different local map representations used for planning. a) The polygons of the global map~$\mathbf{M}$ are converted to a static grid $\mathbf{S}$ shown in orange. The red area is the dilated and blurred static grid $\mathbf{S}'$. The extracted static polygons $\mathbf{P}$ are indicated by a black outline. b) Dynamic polygons have been extracted from the sensed point cloud and expanded for path planning afterwards. c) The dynamic polygonal map $\mathbf{D}(t)$ used for local path planning when computing the heuristic function of the A* search.}
%      \vspace{-5mm}
%\label{fig:pipeline}
%\end{figure*}

Our motion control framework is targeted to control non-holonomic vehicles. It computes acceleration commands at a high control rate using our Short-Term Aborting A*~(STAA*) motion planner. Our framework involves a number of components such as the environment representation, a bounded local planning area, a unicycle model for motion prediction, collision checking, and cost and heuristic functions that determine the order of the STAA* node expansion. In the following, we discuss these components in detail.

\subsection{Prerequisites}

Our motion controller is placed in a larger robot operating system where a global polygonal map $\mathbf{M}$ and a pose estimate~$S$ in the map are available. The polygonal map~$\mathbf{M}$ can be composed of non-convex polygons of arbitrary shape that are allowed to overlap. Additionally, we assume we can observe dynamic polygons $\mathbf{D}$ corresponding to moving obstacles in the sensory field of the controlled agent. These polygons are tracked over multiple frames so that their velocities can be determined. Since moving objects are typically compact, it makes sense to model them as convex polygons as this has a positive implication on collision checking that we will point out in \secref{chap:collisionchecking}, but they can be non-convex if needed. Finally, we assume the current pose of the robot \mbox{$S = (x_S,y_S,\theta_S)$} and a goal pose~\mbox{$G = (x_G,y_G,\theta_G)$} where $x$ and $y$ are the Cartesian coordinates and $\theta$ is the orientation of the robot.

\subsection{Path Planning}
\label{sec:pathplanning}

\begin{figure}[t]
      \centering
      \frame{\includegraphics[height=4cm]{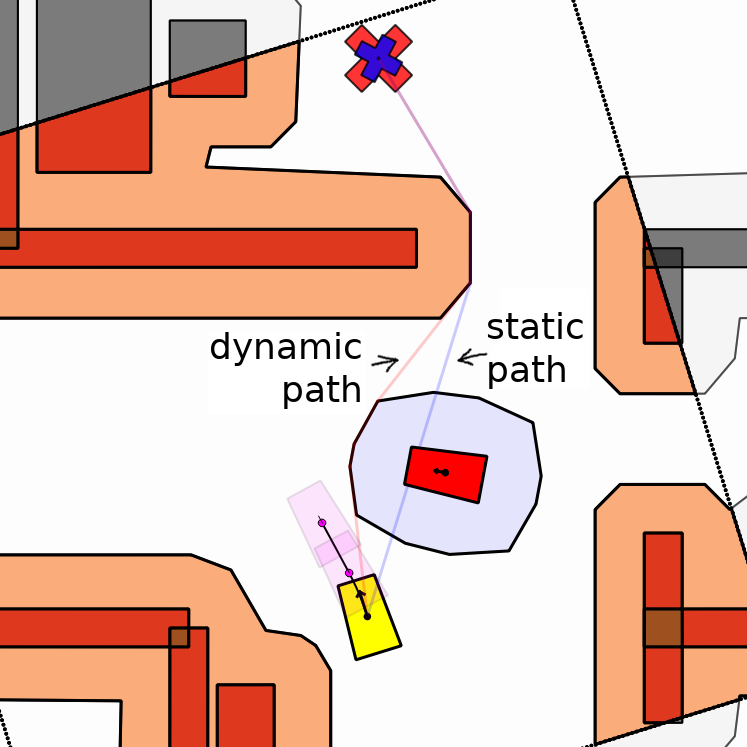}}
      \hfill
	  \frame{\includegraphics[height=4cm]{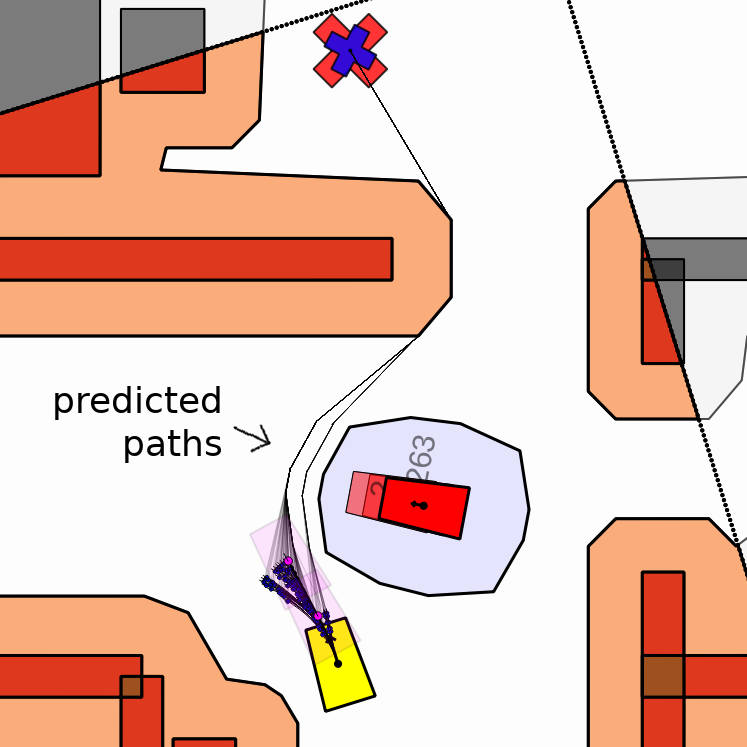}}
	  \caption{Left: A dynamic path that leads around the inflation zone of a dynamic obstacle and a static path that disregards the dynamic obstacle. Right: During the A* planning, the inflation zones of predicted states of the dynamic obstacles are taken into account for shortest path planning.} 
      \label{fig:paths}
\vspace{-6mm}
\end{figure}

For the computation of the global path $P(S,G) = \lbrace(x_i,y_i) \mid i = 0..k\rbrace$ where $(x_0, y_0) = (x_S, y_S)$ and $(x_k, y_k) = (x_G, y_G)$, we use inflated copies $\mathbf{M}'$ and $\mathbf{D}'$ of the polygonal map and the dynamic polygons. The inflation by at least the radius of the agent is applied to avoid planning through too narrow passages the agent would not fit through. Using the Minimal Construct algorithm~\cite{Missura:MinimalConstruct}, we first compute a dynamic global path in $\mathbf{M}' \cup \mathbf{D}'$, the union of the static and the dynamic polygons. If successfully found, the dynamic path helps to guide the agent around moving obstacles. When a moving obstacle blocks a door or a narrow corridor, the dynamic path cannot be found and we fall back to a static path that is computed considering only the static obstacles in $\mathbf{M}'$. \figref{fig:paths} on the left illustrates a dynamic path that has been planned around the inflation zone of a dynamic obstacle. Due to the superior performance of the Minimal Construct algorithm, we can afford to recompute the global path in every control cycle. However, since the completion in one cycle is not guaranteed in extreme large maps, the overall architecture remains two-tiered.

\subsection{Planning in a Local Map}
\label{sec:localmap}
The STAA* motion planner operates in a bounded local map as shown in \figref{fig:everything}. In our implementation, the local map is an 8m$\times$8m square that extends four meters to the left and right, six meters to the front, and two meters to the back of the robot. The local map bounds the complexity of the environment and has a strong impact on the computation times that can be achieved. Motion planning including shortest path computations during the node expansion of A* take place only within the local map towards an intermediate goal pose~$\tilde{G}$. Outside of the local map, the global path $P(S,G)$ completes the plan to the target. 

We determine the intermediate goal pose~$\tilde{G}$ by intersecting the global path with the local map, i.e., we follow $P(S,G)$ from the start $S$ towards the goal $G$ until the first intersection with the boundary of the local map is found, or the goal is reached. Thus, the intermediate goal~$\tilde{G}$ is located either on the boundary of the local map, or inside the local map if the entire path lies inside in which case $\tilde{G} = G$. 

\subsection{Local Map Representations}

The local environmental model consists of a set of grid and geometric maps used for either collision checking or local path planning. Using this hybrid representation, we can exploit the advantages of both types of structures.

The grid representation of the local map is an occupancy grid with a managable size of 160x160 cells and a resolution of 5\,cm. We refer to this grid as the collision grid $\mathbf{C}$ that we compute by drawing the polygons of the global map $\mathbf{M}$ onto the grid. $\mathbf{C}$ is used as a lookup table for fast collision checking during the STAA* search~(cf. \secref{chap:collisionchecking}). Then, we dilate and Gaussian blur $\mathbf{C}$ to an inflated and smoothed costmap $\mathbf{C}'$ to be used as a cost component for the STAA*~(\eqref{eq:cost}). Finally, we extract from the inflated global map $\mathbf{M}'$ a \textit{local} map of non-convex polygons $\mathbf{L}$ that is used to carry out a large number of shortest path computations for the heuristic of the STAA*~(\secref{chap:heuristic}).

The dynamic obstacles $\mathbf{D}$ also become a part of the local model. Their velocities are used for the computation of predicted future states $\mathbf{D}(t)$ by simply moving the dynamic polygons with a constant velocity for time $t$. The predicted states $\mathbf{D}(t)$ are used for polygon vs polygon collision checking by the STAA* (\secref{chap:collisionchecking}) and the predicted and inflated dynamic polygons $\mathbf{D}'(t)$ are used for shortest path planning inside the STAA* heuristic (\secref{chap:heuristic}). See \figref{fig:everything} for an illustration of the concepts described above.

\subsection{Short-Term Aborting A*}

The centerpiece of our framework is a traditional A* search, even though it has been specifically tailored for early abortion. The input of the STAA* is the start pose~$S$, the intermediate goal pose $\tilde{G}$, the collision grid $\mathbf{C}$, the costmap $\mathbf{C}'$, the local polygon map $\mathbf{L}$, and the dynamic polygons $\mathbf{D}$. The output of the planner are linear and angular accelerations $(a,b)$ for the robot to apply until the next control cycle. Let $s = (x,y,\theta,v,\omega)$ be a unicycle state with coordinates $(x,y)$, orientation $\theta$, linear velocity $v$, and angular velocity $\omega$. The initial unicycle state of the search $s_0 = (x_S,y_S,\theta_S, v_S, \omega_S)$ lies in the origin of the local map and contains the sensed velocities $v_S$ and $\omega_S$. The initial state is pushed into the priority queue to initialize the search.

\subsubsection{Action Set and Prediction}

The action set
\begin{equation}
 \mathbb{A} = \left\{ 
 \begin{pmatrix}
 a_{\text{min}}+\frac{i}{n-1}\left(a_{\text{max}}-a_{\text{min}}\right)
  \\ b_{\text{min}}+\frac{j}{n-1}\left(b_{\text{max}}-b_{\text{min}}\right)
 \end{pmatrix}
 \middle| \;
 i,j = 0 .. n-1
 \right\},\label{eq:set}
\end{equation}
of the STAA* consists of tuples~$(a,b)$, where $a$ is the linear acceleration and $b$ is the angular acceleration. The action set is systematically sampled from the permitted ranges $\left[a_{\text{min}}, a_{\text{max}}\right]$ and $\left[b_{\text{min}}, b_{\text{max}}\right]$, respectively. We take $n = 7$ samples from each range and produce an action set of 49 members in total. Given a unicycle state $s_k$, we predict the successor state $s_{k+1} = \mathcal{P}(s_k, (a,b), \bar{t})$ using
\begin{align}
s_{k+1} =
\left(
\begin{array}{l}
  x_k + r\,(\sin(\theta_k + \bar{t} \omega)-\sin(\theta_k)) \\
  y_k - r\,(\cos(\theta_k + \bar{t} \omega)-\cos(\theta_k)) \\  
  \theta_k + \bar{t}\,\omega \\
  v_k + \bar{t}\,a \\
  \omega_k + \bar{t}\,b
\end{array}
\right)
\label{eq:predict}
\end{align}
where $\bar{t} = 0.3$\,s is the prediction time that we keep constant. We determined $n$ and $\bar{t}$ by experimentation and found that those values yield a good trade off between precision and computation time. The future position $(x_{k+1},y_{k+1})$ is computed by an arc approximation according to \cite{Missura:DynamicDWA} similar to \cite{DWA}. The radius $r = \frac{v_k\,+\,0.5\,\bar{t}\,a}{w_k\,+\,0.5\,\bar{t}\,b}$ of the arc is obtained from the linear and angular velocities in the middle of the prediction interval $\bar{t}$. Note that we refrain from using a state lattice, or snapping $s_{k+1}$ to a grid.

\subsubsection{A* Node Expansion and Closing}

The A* algorithm proceeds by popping the topmost state $s_k$ from the priority queue, and applying each acceleration \mbox{$(a,b)_i \in \mathbb{A}$} in the action set to predict new states \mbox{$s_{k+i} = \mathcal{P}(s_k, (a,b)_i)$} according to \eqref{eq:predict}. In the following, we drop the index $i$ and just refer to a successor state $s_{k+1}$. We maintain a table for closing states where we look up the popped state $s_{k}$ and ignore it if its cell is already closed. We close its cell in any case and ignore all subsequent states that fall into a closed cell. The closed table is three-dimensional, coincides with the local grid with a cell size of 5\,cm, and has a resolution of 0.1 radians for the orientation as the third dimension. 

\subsubsection{Collision Checking}
\label{chap:collisionchecking}

The states $s_{k+1}$ predicted by the action set are checked for collisions with static and dynamic obstacles. First, we can quickly discard states where any of the corners of the agent polygon, or its centroid, fall into an occupied cell in the collision grid $\mathbf{C}$. For best results, the agent polygon should be designed with a sufficient number of vertices that are close enough to each other in relation to the resolution of the collision grid. Only for the remaining states that survive the collision table lookup, we perform a polygon vs polygon collision test between the agent polygon at $s_{k+1}$ and the polygons of the dynamic obstacles $\mathbf{D}(t)$ in their predicted states at time \mbox{$t = d(s_{k+1})\,\bar{t}$} where $d(s_{k+1})$ is the depth of $s_{k+1}$ in the search graph. Collided states are discarded and not pushed into the priority queue. Because of this, the queue may run empty when all popped states collide, in which case the agent executes an emergency brake maneuver. The computations of the cost function and the heuristic function are placed after the collision check, because the collision check may discard states for which we do not need to compute the cost and the heuristic.

As long as the agent polygon and the dynamic polygons are convex, polygon vs polygon collision tests can be computed efficiently with the SAT algorithm. If non-convex polygons are required, an approximative collision test can be used that determines whether one polygon contains a corner of the other. This is just as fast as SAT, even though it no longer guarantees to find all collisions. The rare cases it does miss where two polygons overlap but still neither contains a corner of the other, can be avoided by proper design of the hull polygons.

\subsubsection{Cost Function}

After the collision check, we compute $g(s_{k+1})$, the path cost so far, $h(s_{k+1}, \tilde{G})$, the estimated cost-to-go to the intermediate goal $\tilde{G}$, and $f(s_{k+1}) = g(s_{k+1})+h(s_{k+1}, \tilde{G})$, the value the states are ordered by in the priority queue. We express the costs
\begin{equation}
g(s_{k+1}) = t + w_s \mathbf{C}'(s_{k+1}) + w_d \delta(\mathbf{D}(t), s_{k+1}),
\label{eq:cost}
\end{equation}
mostly in terms of the time $ t = d(s_{k+1})\,\bar{t}$ needed to reach the depth $d(s_{k+1})$, but also add the weighted proximity cost obtained from the costmap~$\mathbf{C}'$ at the location of $s_{k+1}$, and the weighted proximity to the dynamic obstacles with \mbox{$\delta(\mathbf{D}(t), s_{k+1}) = \max(1 - |\mathbf{D}(t), s_{k+1}|, 0)$} where $|\mathbf{D}(t), s_{k+1}|$ is the distance to the nearest edge or vertex in the predicted polygonal set $\mathbf{D}(t)$ with a chosen maximum effective proximity range of $1\,$m.

\subsubsection{Heuristic Function}
\label{chap:heuristic}

\begin{figure}[t]
      \centering
      \begin{subfigure}{.39\linewidth}
      \includegraphics[height=2.2cm]{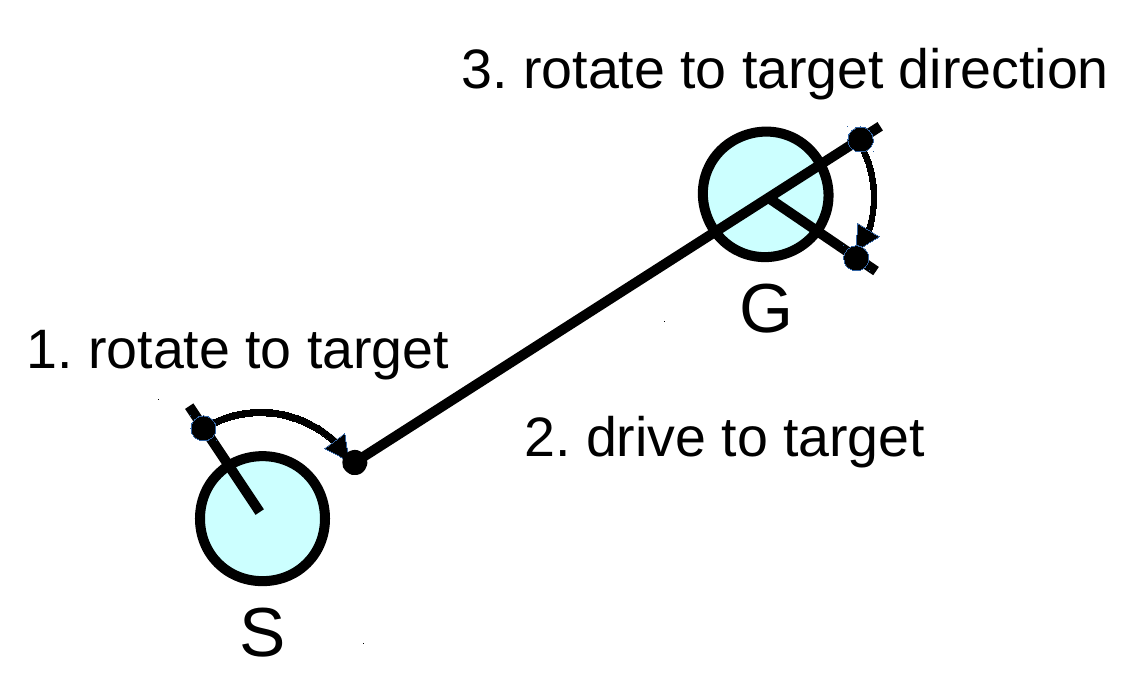}
      \caption{RTR}
      \end{subfigure}
      \hfill
      \begin{subfigure}{.59\linewidth}
      \includegraphics[height=2.2cm]{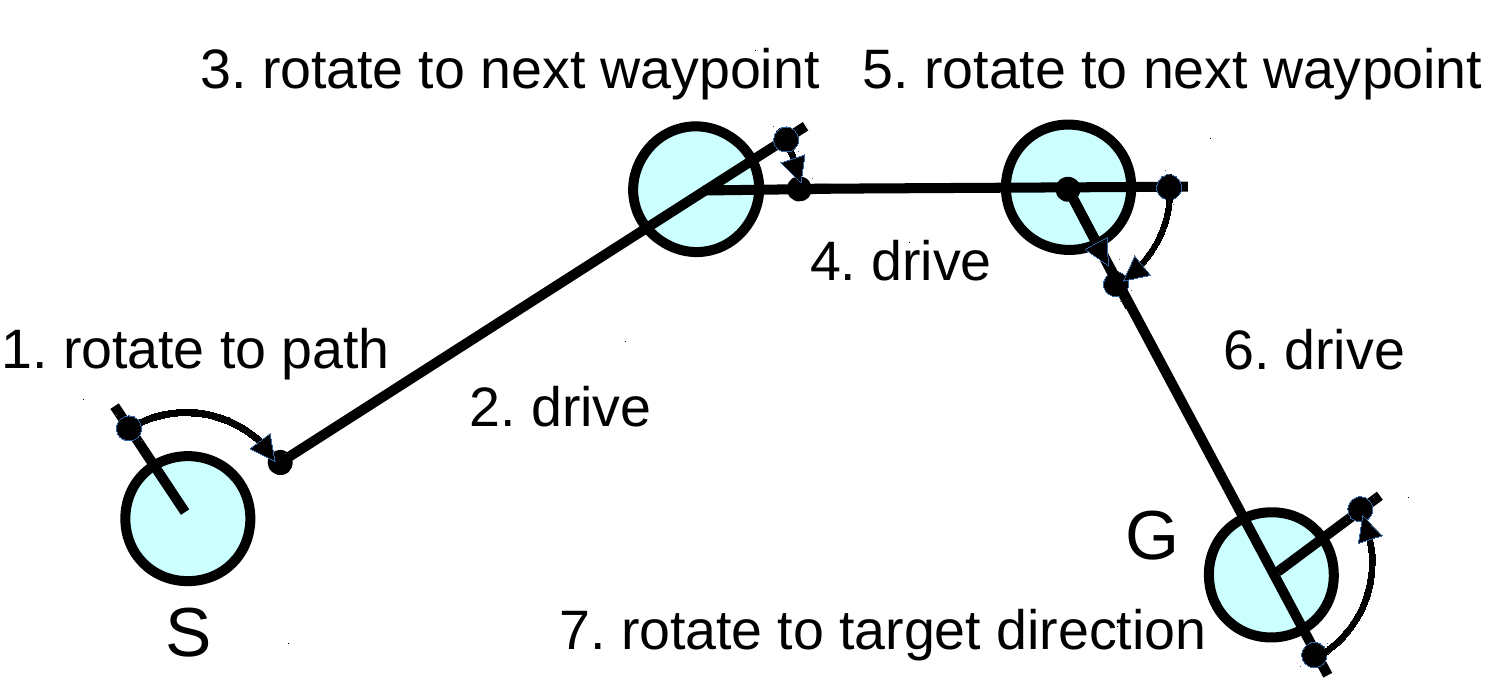}
      \caption{Path RTR}
      \end{subfigure}
      \caption{The rotate-translate-rotate (RTR) heuristic is extended to a Path RTR heuristic by following a path with RTR motions~\cite{Missura:FastFootstepPlanning}.}
      \label{fig:rtr}
\vspace{-5mm}
\end{figure}

For the computation of the heuristic, we determine the shortest path $P(s_{k+1},\tilde{G})$ for each opened state $s_{k+1}$ to the intermediate goal~$\tilde{G}$ according to the same procedure as the global path is found in \mbox{\secref{sec:pathplanning}}, except that now the bounded local map $\mathbf{L}$ can be used instead of the global map $\mathbf{M}$. We first compute the path in $\mathbf{L} \cup \mathbf{D}'(t)$, the union of the static polygons~$\mathbf{L}$ and the predicted and inflated dynamic polygons $\mathbf{D}'(t)$ at time $t$. If no dynamic path can be found, we use a static path obtained only from $\mathbf{L}$. This predicted path planning during the A* search is illustrated in \figref{fig:paths} on the right. Note that we only compute predictions to a depth of 3 as predictions quickly become inaccurate and can do more harm than good. At search depths deeper than 3, we no longer regard $\mathbf{D}(t)$.

Once we have obtained the path for a state $s_{k+1}$, we use the rotate-translate-rotate~(RTR) function to estimate the time needed to drive along a path to the intermediate goal and to attain the goal direction as shown in \figref{fig:rtr}. 
We have successfully applied this heuristic to footstep planning in previous work \cite{Missura:FastFootstepPlanning}. The RTR time function is given by
\begin{equation}
\text{RTR}(s, g) = \frac{|\angle(s, g) - s_{\theta}|}{\omega_{\text{max}}}
+ \frac{|g-s|}{v_{\text{max}}}
+ \frac{|g_{\theta}-\angle(s, g)|}{\omega_{\text{max}}} 
\end{equation}
where $s$ and $g$ are the start state and the goal state, $\angle(s, g)$ denotes the angle of the vector from the start location to the goal location, and $g_{\theta}$ is the desired orientation at the goal. Rotational angles are divided by the maximum angular velocity $\omega_{\text{max}}$ and the driven distance $|g-s|$ is divided by the maximum linear velocity $v_{\text{max}}$ to hopefully underestimate the needed time. In order to compute the RTR time for the shortest path, we simply concatenate the RTR motions along the sections of the path. Let $P(s_{k+1},\tilde{G}) = \lbrace(x_i,y_i, \theta_i) \mid i = 0..n\rbrace$ be a path to the intermediate goal where $P_0 = s_{k+1}$ and $P_n = \tilde{G}$ with orientations $\theta_{i>0} = \angle(P_{i-1}, P_i)$. Then, the estimated time to drive along path $P$ is given by
\begin{equation}
h(s_{k+1}, \tilde{G}) = \sum_{i=0}^{n-1} \text{RTR}( P_i, P_{i+1}).
\end{equation}
The RTR function overestimates the time costs of states that can be reached by a combination of high velocities, \eg $(v_{\text{max}}, \omega_{\text{max}})$, and states close behind the agent that could be reached faster by simply backing up instead of turning. For the latter case, whenever we compute the RTR function, we compute two versions of it considering the forward and the backward direction. Using the smaller of the two results almost entirely mitigates the weak spot of the RTR function. The RTR heuristic helps the A* algorithm to converge quickly towards the goal and is thus well suited for an aborting search where optimality is out of reach.

\subsubsection{Early Aborting}

After we have computed the path cost $g(s_{k+1})$, the heuristic $h(s_{k+1}, \tilde{G})$, and the priority $f(s_{k+1}) = g(s_{k+1})+h(s_{k+1}, \tilde{G})$, the new state $s_{k+1}$ is pushed into the priority queue and the next state is popped for expansion. When popping a state from the queue, we check whether the computation time has run out. If this is the case, STAA* returns the state with the smallest $h$~value found so far as the best solution. We chose the state with the lowest heuristic value, because the last opened state can be anywhere in the search graph even just one level deep, but the state with the lowest~$h$~value is sure to reach closest to the target because of the shortest path affiliation. If a popped state reaches a heuristic value $h(s_k, \tilde{G}) < 0.1$, the search is finished before time and the popped state is returned as a solution from which parent pointers are followed back to the root of the search graph in order to return the first acceleration command.

In average, the STAA* manages to finish in approx. 25\% of all cases within the allowed time. This happens typically when the target is near and inside the local map, or the target is easy to reach without much cornering or narrow spaces. In the other cases, STAA* expands approx. 1000 nodes before abortion and delivers in average plans of a depth of 10 nodes, \ie a three seconds lookahead with $\bar{t} = 0.3$. Since we regard dynamic obstacles only up to a depth of 3, the early abortion has virtually no effect on collision avoidance.

\section{EXPERIMENTS}
\label{chap:experiments}

\subsection{Runtime Analysis}

%\begin{figure}[t]
%\centering
%\includegraphics[width=0.85\columnwidth]{plots/runtimes}
%\caption{Measured runtimes of our control loop. The perception, path planning, and motion control components all exhibit steady runtimes and total at 31\,ms.}
%\label{fig:runtimes}
%\vspace{-17pt}
%\end{figure}

\begin{figure}[t]
\centering
\includegraphics[width=0.85\columnwidth]{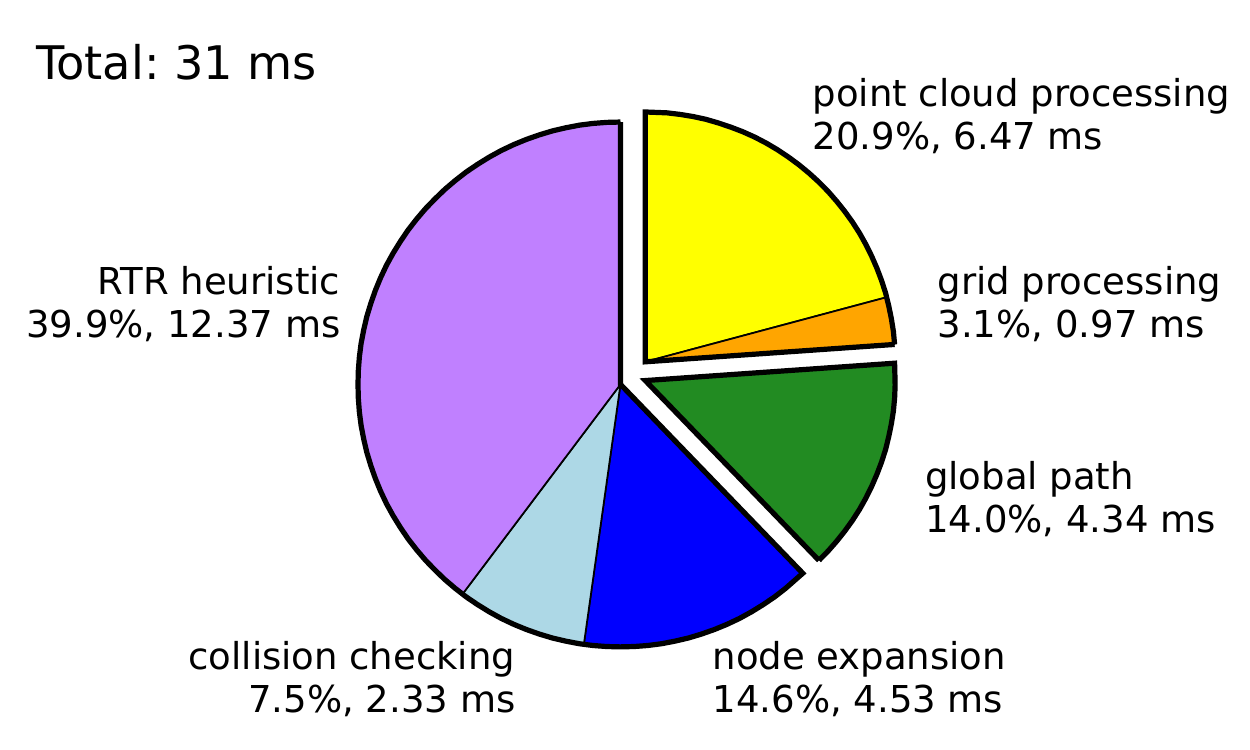}
\caption{Runtime analysis of our motion planning system. The perception pipeline shown in yellow and orange is dominated by sorting the 300K point cloud into an occupancy grid. Postprocessing the grid takes only a fraction of the perception time. The execution time of the STAA* shown in blue shades is taken up mostly by the computation of the heuristic function shown in purple.}
\label{fig:piechart}
\vspace{-6mm}
\end{figure}

\begin{figure*}[t]
      \centering
      \frame{\includegraphics[height=4.5cm]{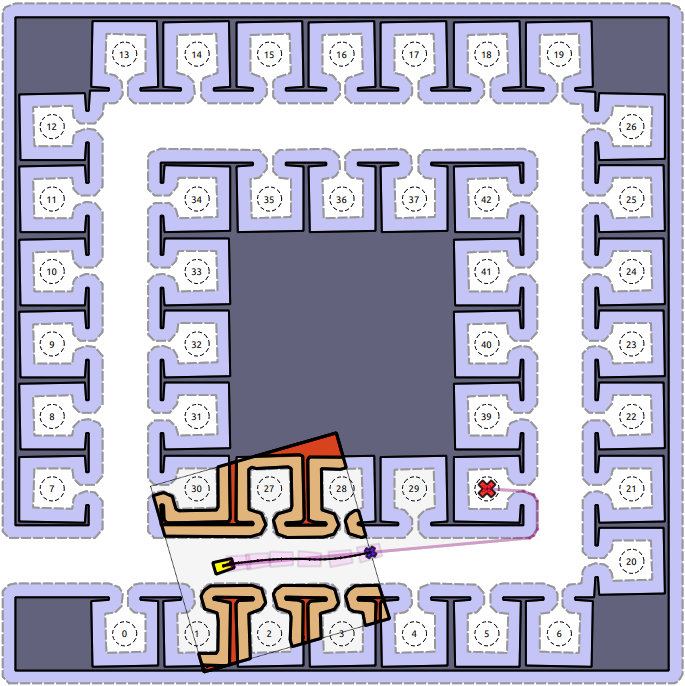}}
      \hspace{1cm}
	  \frame{\includegraphics[height=4.5cm]{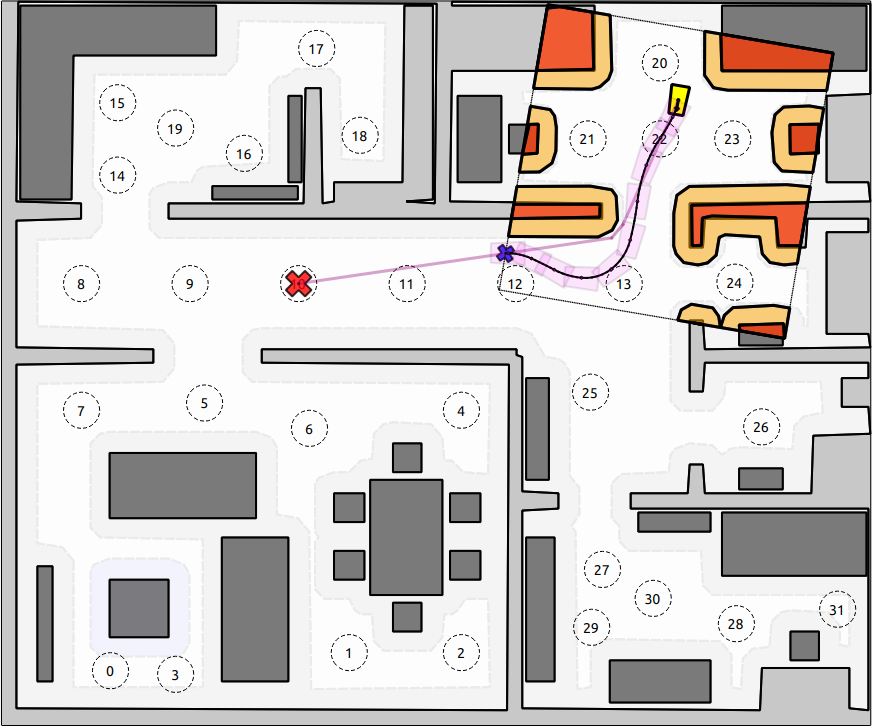}}
	  \hspace{1cm}
      \frame{\includegraphics[height=4.5cm]{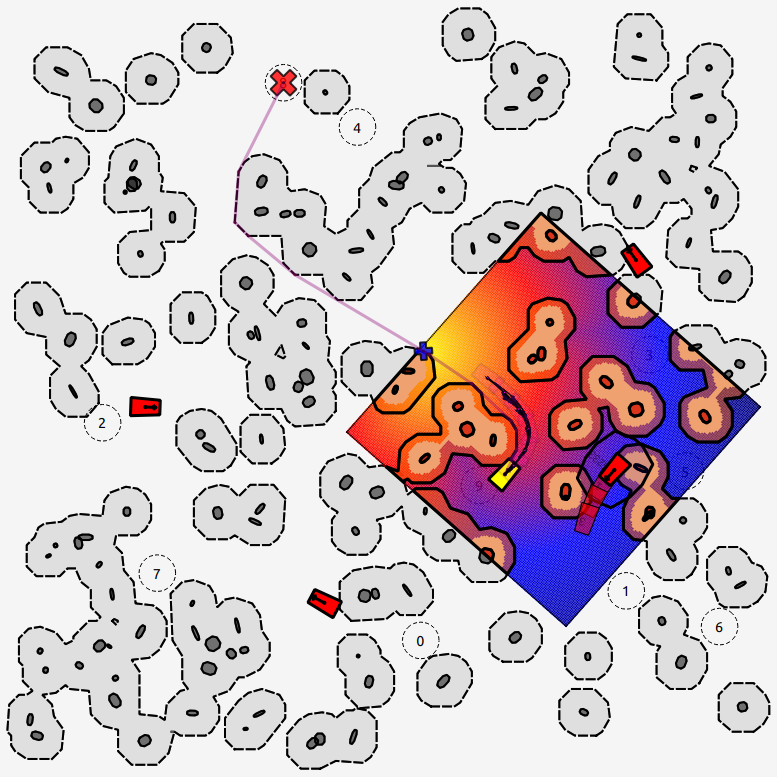}}
      \caption{Simulated environments used for statistical experiments. Left: a 30x30\,m office-like environment (Office). Middle: a 12x10\,m apartment-like map (Apartment). Right: a 20x20\,m clutter scene (Clutter). The circles mark goal locations used for the quantitative evaluation.} 
      \label{fig:maps}
\vspace{-6mm}
\end{figure*}

We profiled our motion control system by running it in combination with our polygonal perception pipeline \cite{Missura:PolygonalPerception} in the same control loop.
The RGB-D sensor delivers data at a rate of 30\,Hz, so this was our target frequency. This leaves us up to 33\,ms to compute an entire cycle from processing a point cloud to computing the acceleration command. \figref{fig:piechart} shows a breakdown of the major computation steps of the control cycle in clockwise order. The perception pipeline takes 7.5\,ms in total. Most of this time is spent on sorting the 300K RGB-D points of the point cloud into the occupancy grid. The grid processing and polygon extraction steps take \mbox{$<$ 1\,ms}. The computation of the global paths takes 4.34\,ms in average. The remaining 21\,ms can be used for the STAA* search. We set the time limit to 19\,ms and average on 31\,ms runtime in total, leaving a little room for fluctuation. %As can be seen in \figref{fig:runtimes}, the control framework manages to uphold a steady frequency.
The runtimes were measured with an Intel® Core \mbox{i7-6800K} 6~x~3.40GHz CPU. 

\subsection{Quantitative Evaluation of Our Controller}

We evaluated the performance of our non-holonomic controller in three simulated environments shown in \figref{fig:maps}. The first environment is an office floor with a size of 30x30\,m. The second environment is the floor plan of an apartment with a size of 12x10\,m. The third environment is a 20x20\,m cluttered scene with small objects with sizes between 3 and 20\,cm, randomly scattered on the floor. In each map, we compared the performance of three different controllers: a simple PD controller, our own DWA implementation with predictive collision avoidance capabilities \cite{Missura:DynamicDWA}, and our new Short-Term Aborting A*. Each controller was tuned to its best performance. The PD controller and the DWA follow a ``carrot'', i.e., a subgoal in a short distance along the global path and benefit from our dynamic path that leads the way around moving obstacles. STAA* aims for the more distant intermediate goal~(\secref{sec:localmap}). For collision avoidance, the PD controller is equipped with a force field that pushes the agent away from obstacles that get too close. The DWA and STAA* use their inbuilt collision avoidance.
The task was to reach as many randomly chosen goal locations (marked with circles in \figref{fig:maps}) as possible within a given time period while also avoiding collisions.

Each controller was tested with up to five agents moving in the same environment in a ``cooperative'' mode, where every agent runs the same controller and makes effort to avoid all collisions, even though the agents do not communicate with each other, and a ghost mode, where only one observed agent is controlled by the evaluated controller and all other agents simply follow their global path with a PD controller completely oblivious of the main agent and each other. In the ghost mode, collisions are counted for the observed agent, but no physical collisions occur and the agents can pass through each other. The ghost mode is more challenging, because the agents are brazenly dashing along their ideal path and provoke unavoidable collisions for the observed agent.

\begin{figure*}[t]
      \centering
	  \includegraphics[height=3.4cm]{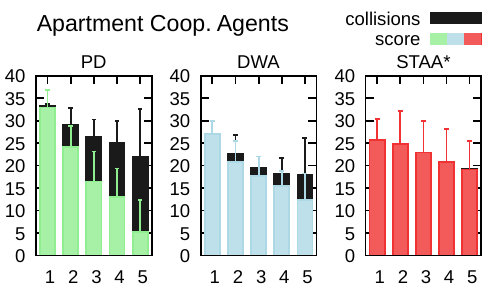}
	  \hfill
      \includegraphics[height=3.4cm]{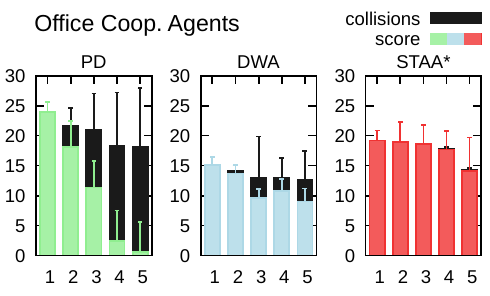}
      \hfill
      \includegraphics[height=3.4cm]{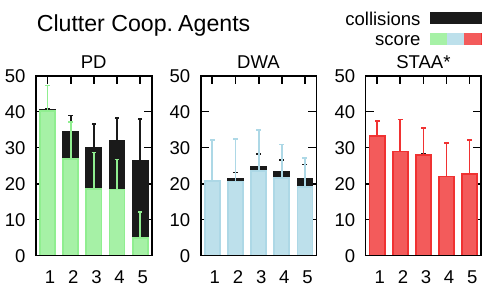}
      
      \includegraphics[height=3.4cm]{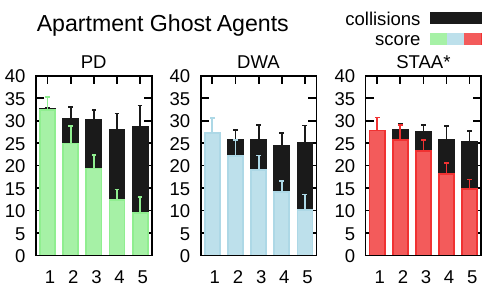}
	  \hfill
      \includegraphics[height=3.4cm]{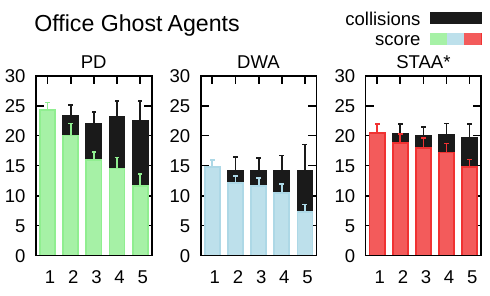}
      \hfill
      \includegraphics[height=3.4cm]{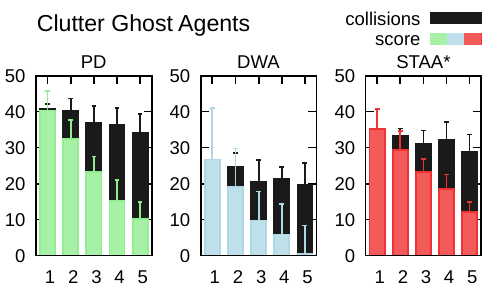}
      
      \caption{Results of the controller evaluation experiment with up to five agents moving in the same environment. Upper row: cooperative agent setting. Bottom row: ghost agent setting. In the cooperative setting, all agents run the same controller, but do not communicate with each other. In the ghost setting, only one observed agent is steered by the evaluated controller, all others blindly follow their shortest path with a PD controller. The colored bars show the score of the respective controller, which is the number of goals reached minus the collisions that occurred. The subtracted collisions are shown in black. As can be seen, the number of collisions produced by our STAA* controller is nearly zero in the cooperative setting and always less than the other controllers in the ghost agent setting. The STAA* achieves a higher score than the DWA in all environments indicating superior performance.}
      \label{fig:fitnessresults}
\vspace{-3mm}
\end{figure*}

The agents were configured with a velocity limit of $2\frac{m}{s}$ in forward direction, $1\frac{m}{s}$ in backward direction, and an angular velocity limit of $3\frac{rad}{s}$. The acceleration limits were set to relatively low values of $2\frac{m}{s^2}$ and $6\frac{rad}{s^2}$, respectively. Each combination of environment, evaluated controller, simulation mode, and number of agents, was ran for 5 minutes and repeated 20~times to generate statistical data. The results are shown in \figref{fig:fitnessresults} where the number of goal locations reached by the observed agent and the number of collisions that the observed agent experienced can be seen. The PD controller drives closely along the shortest path and saturates the maximum forward velocity most of the time, so in terms of reached goals, this simple controller is almost impossible to beat. This comes at the cost of many collisions though, which is highly undesired. Therefore, we evaluate the controllers by a \emph{score} where we subtract the collisions from the number of reached goals. The score is indicated by the colored bars in \figref{fig:fitnessresults} and the amount of collisions that have been subtracted is shown in black. When it comes to avoiding collisions, the force field method of the PD controller is clearly inferior. Both the DWA and the STAA* significantly reduce the number of collisions in all settings compared to the PD controller, even though the STAA* reaches more goals and produces less collisions than the DWA. In the cooperative settings shown in the upper row of \figref{fig:fitnessresults}, the STAA* manages to eliminate collisions almost entirely. Only 4 collisions occurred in total over a simulated driving time of 25 hours. In the ghost setting shown in the bottom row, collisions are unavoidable, but a greater reduction of collisions and a higher score of our STAA* in comparison to the DWA can be observed.

\begin{figure*}[t]
      \centering
	  \includegraphics[height=3.3cm]{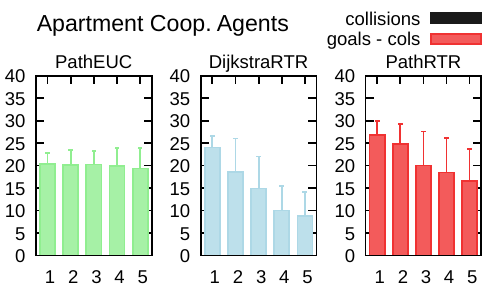}
	  \hfill
      \includegraphics[height=3.3cm]{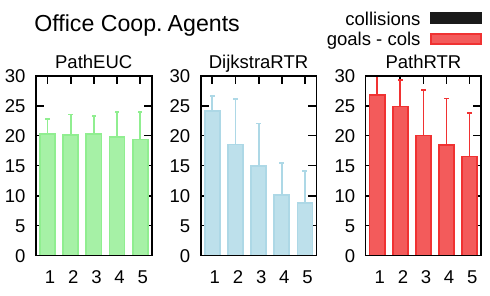}
      \hfill
      \includegraphics[height=3.3cm]{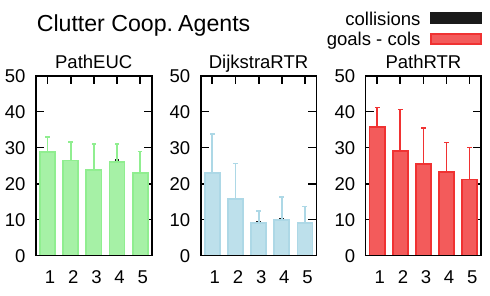}
      
	  \includegraphics[height=3.3cm]{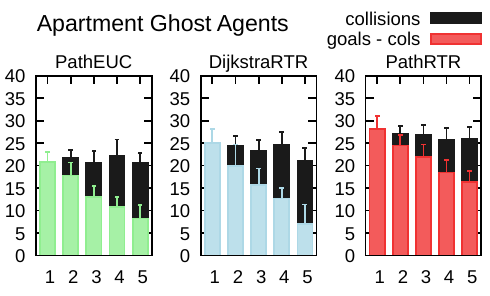}
	  \hfill
      \includegraphics[height=3.3cm]{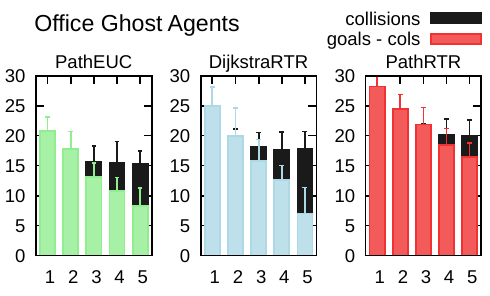}
      \hfill
      \includegraphics[height=3.3cm]{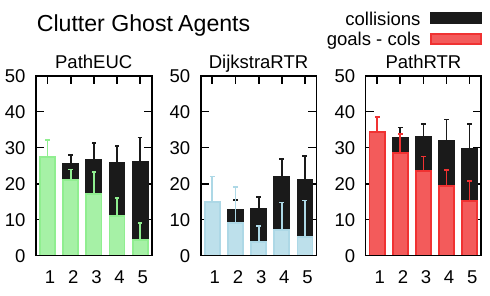}
      \caption{Results of the heuristic experiment with up to five agents moving in the same environment. Upper row: cooperative agent setting. Bottom row: ghost agent setting. In the cooperative setting, all agents run the same controller, but do not communicate with each other. In the ghost setting, only one observed agent is steered by the evaluated controller, all others blindly follow their shortest path with a PD controller. The colored bars show the score of the respective heuristic, which is the number of goals reached minus the collisions that occurred. The subtracted collisions are shown in black. As can be seen, in general the Path RTR heuristic reaches higher scores than the other two candidates. In the ``cooperative'' setting, either of these heuristic functions is sufficient to eliminate almost all collisions with our STAA* controller.}
      \label{fig:heuristicresults}
\vspace{-6mm}
\end{figure*}

\subsection{Evaluation of Different Heuristic Functions}

Furthermore, we evaluated three different heuristic functions to determine their impact on the performance of the STAA*. Apart from our Path RTR heuristic~(\secref{chap:heuristic}), we also considered a Path Euclidean heuristic, where only the length of the shortest path is taken into account but not the changes in orientation, and a Dijkstra heuristic, where a Dijkstra map is computed in the local grid and used as a lookup table for shortest paths which are then used as input for the RTR function. The Dijkstra RTR function is particularly interesting, because after the extra time needed to compute the map, the heuristic function can be evaluated much faster by table lookup. The Dijkstra map is visualized in \figref{fig:maps} in the Clutter map.

The results of the evaluation are shown in \figref{fig:heuristicresults}. Our Path RTR heuristic achieves the best results with higher scores and less collisions than the other candidates. Surprisingly, the Dijkstra RTR heuristic performs worst of the three candidates. This is due to the computation of the Dijkstra table taking up 10\,ms or more of the control cycle. There is not enough time left to profit from the fast lookups and the A* search ends up evaluating fewer options. The Dijkstra table computation is also sensitive to the environment and takes longer on the Clutter map, where the performance of the Dijkstra RTR function falls further back behind the others. The Path Euclidean heuristic performs surprisingly well, even though it ignores the cost of rotation and has a tendency to drive backwards where a lower velocity limit applies. In our simulation it gets away with it due to the omnidirectional sensing, but under realistic circumstances driving backwards is highly undesired. Most interestingly, when all agents are using our STAA* as controller, almost all collisions can be avoided no matter which heuristic is being used.

We also gave Reeds-Shepp curves \cite{ReedsShepp} consideration to be used as heuristic, but determining the Reeds-Shepp curve involves the evaluation of over 40 combinations of left or right turns and straight segments, and this would have to be repeated for each segment of the shortest path. Thus, we ruled this option out due to the considerable computational burden. 

\subsection{Further Experiments}

Furthermore, we repeated the experiments above at replanning rates of 20\,Hz and 10\,Hz, allowing the STAA* a computation time limit of 36 and 84 milliseconds, respectively, and found no noticable decrease or improvement of performance. We also experimented with predicting the trajectories of the controlled vehicle more precisely with Fresnel spirals \cite{Missura:wheeled} instead of the simplistic arc model and found no improvement in performance. Finally, we evaluated a unicycle-based prediction model assuming we can observe not only the linear velocity, but also the angular velocity of other agents and found no significant improvement over the constant velocity model. 

We provide a demonstration video\footnote{\url{http://hrl.uni-bonn.de/publications/staa_simulation.mp4}} of our algorithm performing in simulation, and a video\footnote{\url{http://hrl.uni-bonn.de/publications/staa_realrobot.mp4}} of the Toyota HSR being driven by our STAA* in a real robot experiment.

%%%%%%%%%%%%%%%%%%%%%%%%%%%%%%%%%%%%%%%%%%%%%%%%%%%%%%%%%%%%%%%%%%%%%%%%%%%%%%%%
\section{CONCLUSIONS}

In conclusion, we proposed a fast-paced control framework with an aborting A* motion planner for non-holonomic agents. The control framework uses a geometric map representation and performs collision checks with predicted states of dynamic obstacles. The runtime of the control cycle is bounded so that the system can be used as a low-level replanning controller. The achievement of lower than 33\,ms runtimes is due to the early aborting capability of our A* implementation, which in turn is supported by our Path RTR heuristic function. The bounding of the planning area to a local map also plays a load-bearing role in reducing the runtime. In our simulated experiments, we were able to show a significant decrease of collisions compared to two other non-holonomic controllers. The source code is released here\footnote{\url{https://github.com/MarcellMissura/SpaceTaxi}}.

\bibliographystyle{unsrt}
\bibliography{Bibliography}

\begin{thebibliography}{10}

\bibitem{Missura:DynamicDWA}
Marcell Missura and Maren Bennewitz.
\newblock Predictive collision avoidance for the dynamic window approach.
\newblock In {\em Proc.\ of the {IEEE} Int.\ Conf.\ on Robotics \& Automation
  (ICRA)}, 2019.

\bibitem{DWA}
D.~Fox, W.~Burgard, and S.~Thrun.
\newblock The dynamic window approach to collision avoidance.
\newblock {\em IEEE Robotics Automation Magazine}, Mar 1997.

\bibitem{OfficeMarathon}
E.~Marder-Eppstein, E.~Berger, T.~Foote, B.P. Gerkey, and K.~Konolige.
\newblock The office marathon: {R}obust navigation in an indoor office
  environment.
\newblock In {\em Proc.\ of the {IEEE} Int.\ Conf.\ on Robotics \& Automation
  (ICRA)}, 2010.

\bibitem{Missura:FastFootstepPlanning}
M.~Missura and M.~Bennewitz.
\newblock Fast footstep planning with aborting {A*}.
\newblock In {\em Proc.\ of the {IEEE} Int.\ Conf.\ on Robotics \& Automation
  (ICRA)}. {IEEE}, 2021.

\bibitem{Missura:wheeled}
Marcell Missura and Sven Behnke.
\newblock Efficient kinodynamic trajectory generation for wheeled robots.
\newblock In {\em 2011 IEEE International Conference on Robotics and
  Automation}, 2011.

\bibitem{GlobalDWA}
Oliver Brock and Oussama Khatib.
\newblock High-speed navigation using the global dynamic window approach.
\newblock In {\em In IEEE Int. Conf. on Robotics and Automation (ICRA)}, pages
  341--346, 1999.

\bibitem{SIPP}
Mike Phillips and Maxim Likhachev.
\newblock Sipp: Safe interval path planning for dynamic environments.
\newblock In {\em ICRA}, pages 5628--5635, 2011.

\bibitem{TBAS}
Yngvi Bj{\"{o}}rnsson, Vadim Bulitko, and Nathan~R. Sturtevant.
\newblock Tba*: Time-bounded a*.
\newblock In Craig Boutilier, editor, {\em {IJCAI} 2009, Proceedings of the
  21st International Joint Conference on Artificial Intelligence, Pasadena,
  California, USA, July 11-17, 2009}, pages 431--436, 2009.

\bibitem{RTAA}
Sven Koenig and Maxim Likhachev.
\newblock Real-time adaptive a*.
\newblock In Hideyuki Nakashima, Michael~P. Wellman, Gerhard Weiss, and Peter
  Stone, editors, {\em 5th International Joint Conference on Autonomous Agents
  and Multiagent Systems {(AAMAS} 2006), Hakodate, Japan, May 8-12, 2006},
  pages 281--288. {ACM}, 2006.

\bibitem{LRTAS}
Richard~E. Korf.
\newblock Real-time heuristic search.
\newblock {\em Artif. Intell.}, 42(2–3):189–211, mar 1990.

\bibitem{DStarLite}
S.~Koenig and M.~Likhachev.
\newblock ${D}^*$ {L}ite.
\newblock In {\em Proc.~of the National Conference on Artificial Intelligence
  (AAAI)}, 2002.

\bibitem{Likachev}
Maxim Likhachev and Dave Ferguson.
\newblock Planning long dynamically feasible maneuvers for autonomous vehicles.
\newblock {\em The International Journal of Robotics Research}, 28(8):933--945,
  2009.

\bibitem{ARAStar}
M.~Likhachev, D.~Ferguson, G.~Gordon, A.~Stentz, and S.~Thrun.
\newblock Anytime search in dynamic graphs.
\newblock {\em Artificial Intelligence}, 172(14):1613 -- 1643, 2008.

\bibitem{Lau}
B.~Lau, C.~Sprunk, and W.~Burgard.
\newblock Kinodynamic motion planning for mobile robots using splines.
\newblock In {\em Proc.\ of the {IEEE} Int.\ Conf.\ on Intelligent Robots \&
  Systems (IROS)}, 2009.

\bibitem{Usenko}
Vladyslav~C. Usenko, Lukas von Stumberg, Andrej Pangercic, and Daniel Cremers.
\newblock Real-time trajectory replanning for mavs using uniform b-splines and
  3d circular buffer.
\newblock {\em CoRR}, abs/1703.01416, 2017.

\bibitem{Sterling}
Sterling McLeod and Jing Xiao.
\newblock Real-time adaptive non-holonomic motion planning in unforeseen
  dynamic environments.
\newblock In {\em {IROS}}, pages 4692--4699. {IEEE}, 2016.

\bibitem{Kinodynamic}
C.~R{\"{o}}smann, F.~Hoffmann, and T.~Bertram.
\newblock Kinodynamic trajectory optimization and control for car-like robots.
\newblock In {\em Proc.\ of the {IEEE} Int.\ Conf.\ on Intelligent Robots \&
  Systems (IROS)}, 2017.

\bibitem{Siegwart}
Mark Pfeiffer, Michael Schaeuble, Juan~I. Nieto, Roland Siegwart, and Cesar
  Cadena.
\newblock From perception to decision: {A} data-driven approach to end-to-end
  motion planning for autonomous ground robots.
\newblock {\em CoRR}, abs/1609.07910, 2016.

\bibitem{Everett1}
Yu~Fan Chen, Michael Everett, Miao Liu, and Jonathan~P. How.
\newblock Socially aware motion planning with deep reinforcement learning.
\newblock In {\em {IROS}}, pages 1343--1350. {IEEE}, 2017.

\bibitem{Everett2}
Michael Everett, Yu~Fan Chen, and Jonathan~P. How.
\newblock Motion planning among dynamic, decision-making agents with deep
  reinforcement learning.
\newblock {\em CoRR}, abs/1805.01956, 2018.

\bibitem{Kretzschmar}
Henrik Kretzschmar, Markus Spies, Christoph Sprunk, and Wolfram Burgard.
\newblock Socially compliant mobile robot navigation via inverse reinforcement
  learning.
\newblock {\em I. J. Robotics Res.}, 35(11):1289--1307, 2016.

\bibitem{VO}
Paolo Fiorini and Zvi Shiller.
\newblock Motion planning in dynamic environments using velocity obstacles.
\newblock {\em The International Journal of Robotics Research}, 17(7):760--772,
  1998.

\bibitem{RVO}
J.~van~den Berg, M.~Lin, and D.~Manocha.
\newblock Reciprocal velocity obstacles for real-time multi-agent navigation.
\newblock In {\em 2008 IEEE International Conference on Robotics and
  Automation}, pages 1928--1935, 2008.

\bibitem{NRVO}
J.~van~den Berg, S.~J. Guy, M.~Lin, and D.~Manocha.
\newblock Reciprocal n-body collision avoidance.
\newblock In {\em Robotics Research}, pages 3--19, Berlin, Heidelberg, 2011.
  Springer Berlin Heidelberg.

\bibitem{AVO}
J.~van~den Berg, J.~Snape, S.~J. Guy, and D.~Manocha.
\newblock Reciprocal collision avoidance with acceleration-velocity obstacles.
\newblock In {\em Proc.\ of the {IEEE} Int.\ Conf.\ on Robotics \& Automation
  (ICRA)}, 2011.

\bibitem{VO3}
J.~Snape, J.~van~den Berg, S.~J. Guy, and D.~Manocha.
\newblock Smooth and collision-free navigation for multiple robots under
  differential-drive constraints.
\newblock In {\em Proc.\ of the {IEEE} Int.\ Conf.\ on Intelligent Robots \&
  Systems (IROS)}, pages 4584--4589. {IEEE}, 2010.

\bibitem{Missura:MinimalConstruct}
Marcell Missura, Daniel~D. Lee, and Maren Bennewitz.
\newblock Minimal construct: Efficient shortest path finding for mobile robots
  in polygonal maps.
\newblock In {\em Proc.\ of the {IEEE} Int.\ Conf.\ on Intelligent Robots \&
  Systems (IROS)}, 2018.

\bibitem{Missura:PolygonalPerception}
Marcell Missura, Arindam Roychoudhury, and Maren Bennewitz.
\newblock Polygonal perception for mobile robots.
\newblock In {\em Proc.\ of the {IEEE} Int.\ Conf.\ on Intelligent Robots \&
  Systems (IROS)}, 2020.

\bibitem{ReedsShepp}
J.~A. Reeds and L.~A. Shepp.
\newblock {Optimal paths for a car that goes both forwards and backwards.}
\newblock {\em Pacific Journal of Mathematics}, 145(2):367 -- 393, 1990.

\end{thebibliography}
\end{document}